\documentclass[9pt, conference]{IEEEtran}
\IEEEoverridecommandlockouts
\usepackage{cite}
\usepackage{amsmath,amssymb,amsfonts}
\usepackage{algorithmic}
\usepackage{graphicx}
\usepackage{textcomp}
\usepackage{xcolor}

\usepackage{multirow}
\usepackage{booktabs}
\usepackage[ruled,linesnumbered]{algorithm2e}
\usepackage[normalem]{ulem}
\usepackage{balance}
\usepackage{threeparttable}

\usepackage[switch]{lineno}

\def\BibTeX{{\rm B\kern-.05em{\sc i\kern-.025em b}\kern-.08em
    T\kern-.1667em\lower.7ex\hbox{E}\kern-.125emX}}
\begin{document}

\title{Testability-Aware Low Power Controller Design with Evolutionary Learning}

\author{\IEEEauthorblockN{Min Li\IEEEauthorrefmark{1}, Zhengyuan Shi\IEEEauthorrefmark{1}, Zezhong Wang\IEEEauthorrefmark{2}, Weiwei Zhang\IEEEauthorrefmark{2}, Yu Huang\IEEEauthorrefmark{2} and Qiang Xu\IEEEauthorrefmark{1}}
\IEEEauthorblockA{\IEEEauthorrefmark{1}\textit{Department of Computer Science and Engineering},
\textit{The Chinese University of Hong Kong},
Shatin, Hong Kong S.A.R.\\
\{mli, zyshi21, qxu\}@cse.cuhk.edu.hk}

\IEEEauthorblockA{\IEEEauthorrefmark{2}\textit{Huawei Technologies Co., Ltd.}, China \\
\{wangzezhong7, zhangweiwei4, huangyu61\}@hisilicon.com
}
}


\maketitle

\begin{abstract}
XORNet-based low power controller is a popular technique to reduce circuit transitions in scan-based testing. However, existing solutions construct the XORNet evenly for scan chain control, and it may result in sub-optimal solutions without any design guidance. In this paper, we propose a novel testability-aware low power controller with evolutionary learning. The XORNet generated from the proposed genetic algorithm (GA) enables adaptive control for scan chains according to their usages, thereby significantly improving XORNet encoding capacity, reducing the number of failure cases with ATPG and decreasing test data volume. 
Experimental results indicate that under the same control bits, our GA-guided XORNet design can improve the fault coverage by up to 2.11\%. 
The proposed GA-guided XORNets also allows reducing the number of control bits, and the total testing time decreases by 20.78\% on average and up to 47.09\% compared to the existing design without sacrificing test coverage.

\end{abstract}


\section{Introduction}

The scale and complexity of integrated circuits (IC) continue to grow thanks to the ever advanced semiconductor technologies. However, the associated large volume of test data to retain satisfactory test coverage leads to significant test cost. To mitigate this issue, a large amount of test compression techniques were proposed in the literature~\cite{rajski2002embedded, rajski2004embedded, chandra2001frequency, koneman1993lfsr, chandra2000test, liang2002two, gonciari2003variable, hellebrand2001mixed, dorsch2001tailoring, das2000reducing, bayraktaroglu2001test}. Nowadays, the mainstream technique is to store the compressed test patterns on the automated test equipment (ATE) and expand them on-chip before loading onto scan chains. 

At the same time, with the continuous downscaling of transistor size, the circuit in scan test mode may behave quite differently from that in functional mode. In particular, scan test patterns could induce much higher switching activities than functional patterns, leading to unnecessary test yield loss.  
To address this critical problem, it is essential to consider the problem of test data compression and the problem of low power test together. Consequently, various \textit{low power test data compression} solutions were proposed~\cite{bhatia2010low, czysz2008low, lee2004low, moghaddam2010low, rosinger2002low, sowmiya2013design, cho2007california, dabholkar1998techniques, bonhomme2003efficient, bonhomme2001gated, huang2001reduction}. Among these works, the design of \textit{low power test controller} has been widely adopted by commercial tools. 
The basic idea is to use a control block to output gating (masking) signals to scan chains such that they can be either connected with the decompressor or fed by a constant value of $0$ ($1$) on a per pattern basis~\cite{czysz2008low}. 

Both test compression and low power test control techniques exploit the fact that scan test patterns often feature a large portion of unspecified bits. Consequently, we could use a \textit{decoder} (instantiated as LFSR reseeding structure~\cite{lee2004low, rosinger2002low} or XORNet~\cite{czysz2008low, moghaddam2010low}) with a small number of control bits to obtain the gating signals in low power test controller. Take XORNet in~\cite{czysz2008low, moghaddam2010low} as an example (see Fig.~\ref{fig:controller}), the gating signal for a specific scan chain is the XOR-ed output from certain control bits. When we need to drive the scan chains with specified bits from the decompressor, the control bits should be configured in a way that the gating signals for these scan chains are all $1$'s. The remaining gating signals are determined by both the XORNet structure and the other control bits. 

In order to minimize the impact of the linear dependency between different gating signals, the conventional XORNet design will evenly distribute the scan chains to certain XORNet control bits (\textit{e.g.}, $3$)~\cite{mrugalski2000linear, czysz2008low, rajski2004embedded}. 
Such configuration implicitly assumes uniform XOR operand setting and equal encoding capacity of each scan chain. 
In practice, different scan chains in a circuit have distinct encoding requirements and they also vary among circuits. Consequently, the number of control bits is usually set conservatively high to ensure high encoding capability (\textit{e.g.}, $25$ control bits for a circuit with $128$ scan chains), leading to sub-optimal controller design. 

In this paper, we take the above issue into consideration and propose a testability-aware low power controller design with \textit{evolutionary learning} techniques. The proposed method implicitly extracts testability information via applying \textit{genetic algorithm (GA)} on the sampled test cubes, and customizes XORNet-based low power controllers for different circuits under test. The proposed GA-guided XORNet design outperforms existing solutions by a significant margin across different settings.

The contributions of this paper are summarized as follows:
\begin{itemize}
    \item We use \textit{fault sampling} to generate sampled test cubes and extract the scan chain usage distribution that embeds testability information.
    \item We propose to apply the genetic algorithm on sampled test cubes to explore a better XORNet structure for the low power test controller design. The genetic algorithm operations are designed specifically for XORNet structures.
    \item We apply GA-guided XORNet design as the low power test controller for various industrial circuits and our experimental results show significant improvements in terms of test coverage, test time and control bit count.
\end{itemize}

We organize the remainder of this paper as follows. We review related works in Section~\ref{sec:related}. Section~\ref{sec:lowpower} introduces the conventional XOR-based low power test controller design, which motivates the proposed genetic algorithm for XORNet designs presented in Section~\ref{sec:ga}. Then, Section~\ref{sec:exp} presents our experimental results on various industrial circuits. Finally, Section~\ref{sec:conclusion} concludes this paper.


\section{Related Works}

In this work, we focus on scan shift power reduction and we discuss previous work in Section~\ref{subsec:lp}. Then, Section~\ref{subsec:el} introduces preliminaries of evolutionary learning algorithms. 
\label{sec:related}
\subsection{Low-Power Test Data Compression}
\label{subsec:lp}
Generally speaking, low shift-power testing techniques can be categorized into \textit{ATPG}-based and \textit{DFT}-based solutions. 

\begin{figure}[t]
    \centering
    \includegraphics[width=0.9\linewidth]{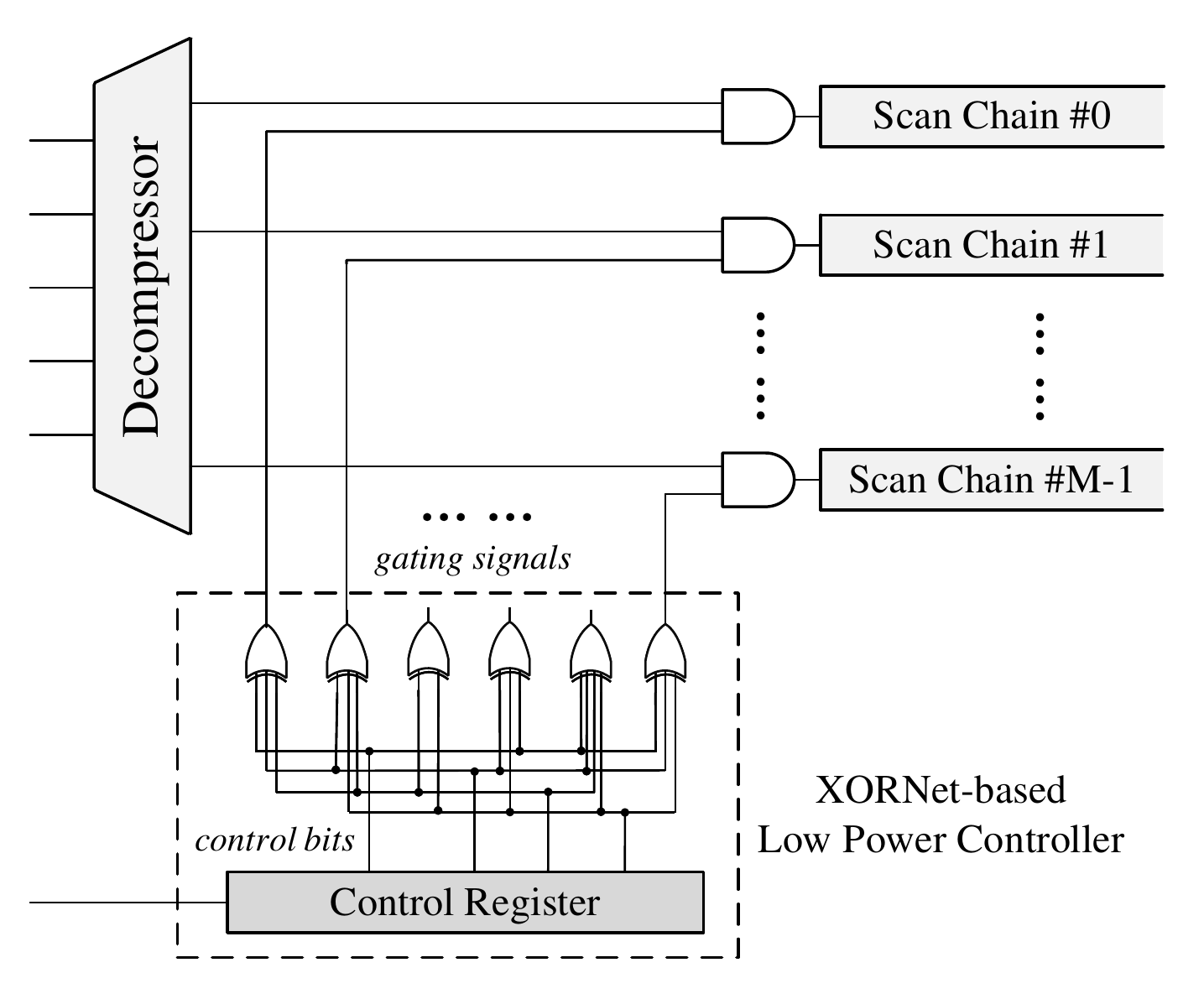}
    \caption{The XORNet-based Low Power Test Compression Architecture~\cite{czysz2008low}}
    \label{fig:controller}
\end{figure}

ATPG-based techniques (\textit{e.g.}, test vector reordering~\cite{dabholkar1998techniques}, adjacent filling~\cite{butler2004minimizing}, X-filling ATPG~\cite{wu2007efficient, wen2006highly}) try to minimize test patterns' switching activities during test generation. However, these techniques are not friendly with test compression and hence are rarely used for shift power reduction in practice.
DFT-based low power test solutions, on the other hand, add design-for-test (DFT) hardware to reduce circuit switching activities during testing. Representative techniques include scan chain partitioning and modification~\cite{rosinger2004scan, bonhomme2003efficient}, clock gating~\cite{bonhomme2001gated}, test point insertion~\cite{elshoukry2005partial}, adaptive power scaling~\cite{devanathan2007pmscan}, and low power controller design~\cite{lee2004low, czysz2008low}. Among them, low power controller can be easily integrated into the test compression environment and is widely adopted in the industry. To be specific, the low power test controller is usually an LFSR-based or XORNet-based encoder, which expands the control bits into gating signals for every scan chain. The gating signals are AND-ed or OR-ed with the decompressed scan data for test power reduction.
One example XORNet-based low power controller design is illustrated in Fig.~\ref{fig:controller}. In this paper, we consider the above low power test compression architecture and try to 
improve its efficiency.

\subsection{Evolutionary Learning Algorithm}
\label{subsec:el}
Inspired by biological evolution, the common idea behind variants of evolutionary learning algorithms~\cite{angeline1994evolutionary, deb2002computationally, eiben2015evolutionary, cohoon2003evolutionary} is the same: given a population of individuals within some environment that has limited resources, competition for those resources causes natural selection. Among different evolutionary learning algorithms, genetic algorithm~\cite{patrascu2014helga, eiben1994genetic, ting2005mean} is arguably the most popular one, which has been widely used for many optimization problems. Given an optimization problem, a population of candidate solutions is evolved towards better solutions. The solutions are in the form of strings of numbers, which is similar to \textit{chromosomes} in genes, and can be mutated and altered. In order to evaluate the performance of different solutions, the \textit{fitness} of the population is defined and serves as the optimization objective. 
The evolution usually starts from a population of randomly generated individuals and then steps into an iterative process. In each generation, the fitness function is applied to every individual; Based on these fitness values, potentially better candidates are stochastically chosen as the mating parents to seed the next generation. The recombination (\textit{i.e.}, \textit{crossover}) and possibly randomly \textit{mutation} are performed to produce new candidates. The new generation of candidate solutions is then used in the next iteration of the algorithm. This process is repeated until a candidate with sufficient quality is found or a previously set computational limit is reached.

\section{XORNet-Based Low Power Controller Design}
\label{sec:lowpower}

The binary vector $\mathbf{x} \in \{0, 1\}^N$ of a test cube, wherein the $i^{th}$ dimension indicates the $i^{th}$ scan chain being \textit{specified} (value 1) or not (value 0), represents the usage of scan chains. We use a XORNet to encode $\mathbf{x}$ into compressed representation $\mathbf{z}$ as the control bits, which will be loaded into the control register before decompression (See Fig.~\ref{fig:controller}). 
The XORNet is a binary matrix, $\mathbf{A} \in \{0, 1\}^{N\times M}$, defined by $\mathbf{A}(i, j) = \mathbb{I}(z_j \in x_i)$ (the notation $z_j \in x_i$ here means that the $j^{th}$ control bit is the operand of XOR for the $i^{th}$ scan chain). The row vector $\mathbf{a}_i$ denotes the operands for the $i^{th}$ scan chain. The control bits $\mathbf{z} \in \{0, 1\}^M$ represents the controlling signal for XORNet, while the expanded scan chain gating signal $\mathbf{\hat{x}} \in \{0, 1\}^N$ is the decoded signal from XORNet. Here we have the following step to get the control signals for test cubes:
\begin{equation}
    \mathbf{z} = \left\{
    \begin{array}{cc}
        \text{XORSolver}(\mathbf{A}, \mathbf{x}), &  \text{if }\mathbf{x} \text{ is solvable} \\
        None, & \text{otherwise}.
    \end{array} \right. 
\end{equation}

The XORSolver uses the \textit{Gauss-Jordan} method, except that we consider the finite field of integers modulo $2$. Note that we only solve the equations with the output being $1$. What's more, when the system has more than one solution, the unspecified control bits are random-filled. On the other hand, if the system has no solution, it counts as a failure case.

If the system is solvable, we can decode the control signals by XOR $\mathbf{A}$ and $\mathbf{z}$, and get the gating signal $\mathbf{\hat{x}}$, wherein the $i^{th}$ entry with value $1$ indicates the $i^{th}$ scan chain is driven by decompressor (\textit{i.e.}, the scan chain is \textit{activated/enabled}):
\begin{equation}
    \mathbf{\hat{x}} = \mathbf{A} \oplus \mathbf{z}
\end{equation}


The design goal of the XORNet-based low power controller is to minimize the power consumption as much as possible, within the minor loss of fault coverage. Usually, the power dissipation should meet the pre-defined requirements so that the overheating problem can be avoided during testing. To model the problem, we formally define a few indicators as follows.

Firstly, during ATPG, XORNet may fail to encode the test cubes. Specifically, the failure cases happen when the system $(\mathbf{A}, x)$ has no solution. We define \textit{Unsolvable} (UNS) as the number of failure cases of XORNet encoding for a given test cube set:
\begin{equation}
    UNS = \# \{\text{XORNet Encoding Failure Case}\}
    \label{eq:UNS}
\end{equation}
It should be noted that when all gating signals are decoded by $2k+1$ ($k$ is an integer) of control bits, there is no UNS. The reason is that when all control bits are set with $1$, the XOR outputs of $2k+1$ ones must be $1$, causing all scan chains are activated. 

Then we consider the cases where the power consumption exceeds the user-defined limit. Toggling the internal value of scan cell directly consumes power in scan testing. Because the excessive power during testing will cause circuit damaged, the ATPG process should specify a peak value as maximum instantaneous power consumption. To measure the shift cycle power in load and unload phases, we follow the definition of scan cell transition rate in \cite{tsai2010test}, denoted as the \textit{Transition Rate} at clock cycle $t$: 
\begin{equation}
    r_t = \frac{\# \{\text{Transition Cells}\} \text{ at } t}{\# \{\text{Cells}\}} \label{eq:transition}
\end{equation}
For example, one scan chain consists of 5 scan cells with internal states $S=\{c_0, c_1, c_2, c_3, c_4\}$ and we assume the current internal states $S_t = \{0, 0, 1, 1, 0\}$. In the next clock cycle, scan cell $c_4$ is loaded $1$, the internal states should be $S_{t+1} = \{0, 1, 1, 0, 1\}$. The scan cells $c_1$ and $c_4$ take falling transition from $1$ to $0$, the scan cells $c_3$ takes rising transition from $0$ to $1$, and other cells keep the state in the last clock cycle. Therefore, there are $3$ transition cells among total $5$ cells, the transition rate at clock $t$ is $r_t = 0.6$. The upper limit of $r_t$ is defined as \textit{Maximum Transition Rate} ($R$), which is determined by the user. If the transition rate of a test cube at any cycle is larger than $R$, this test cube is counted as a \textit{High Power} (HP) cube. The loss of test coverage is caused by discarding such HP cubes. 

Evaluating transition rate for each test cube for all clock cycles is time-consuming. 
To efficiently estimate the power consumption, 
we use the ratio of the activated scan chains to the total number of scan chains to approximate the power consumption, named as \textit{Scan Chain Activated Rate} (SCA). 
\begin{equation}
    SCA(\mathbf{x}) = \frac{\text{\# Activated Scan Chain}}{\text{\# Scan Chain}} \in [0, 1]
    \label{eq:power}
\end{equation}

We then define \textit{Scan Chain Activated Exceeding Counts} (SCAE) as the number of test cubes which exceeding the SCA limit, and the $\overline{SCA}$ as the average scan chain activated rate for a given test cube set.
\begin{equation}
    SCAE = \# \{\mathbb{I}(SCA(\mathbf{x}) > Limit)\}
    \label{eq:powe}
\end{equation}
\begin{equation}
    \overline{SCA} = mean(\{SCA(\mathbf{x})\})
    \label{eq:powermean}
\end{equation}

When a test cube activates too many scan chains or it cannot be encoded by the XORNet, the test cube is regarded as \textit{Unsuccessfully Encoding} (UE) cube. The number of UE is the sum of SCAE and UNS, which can be calculated as:
\begin{equation}
    UE = SCAE + UNS
    \label{eq:UE}
\end{equation}

Not double to say, the primary objective of XORNet-based low power controller is to reduce the cases which violate the power requirements as much as possible. 
However, the conventional XORNet design is merely random and conservative. For example, given a circuit design with $N$ scan chains, $M$ ($M < N$) control bits will be assigned for XORNet, and $3$ control bits are connected randomly to one scan chain, \textit{i.e.}:
\begin{equation}
    \mathbf{\hat{x}}_i = \mathbf{z}_{j}\oplus \mathbf{z}_{k}\oplus \mathbf{z}_{p}
\end{equation}
The random seed to generate XORNet is set as the number of scan chains by convention.
Also, different scan chains in a circuit have distinct encoding requirements and they also vary among circuits. So the number of control bits is usually set conservatively high so that the low power controller does not cause high loss of test coverage. For example, $11$ control bits are allocated for a circuit with $46$ scan chains in the industrial tool.

\begin{figure}[t]
    \centering
    \includegraphics[width=0.8\linewidth]{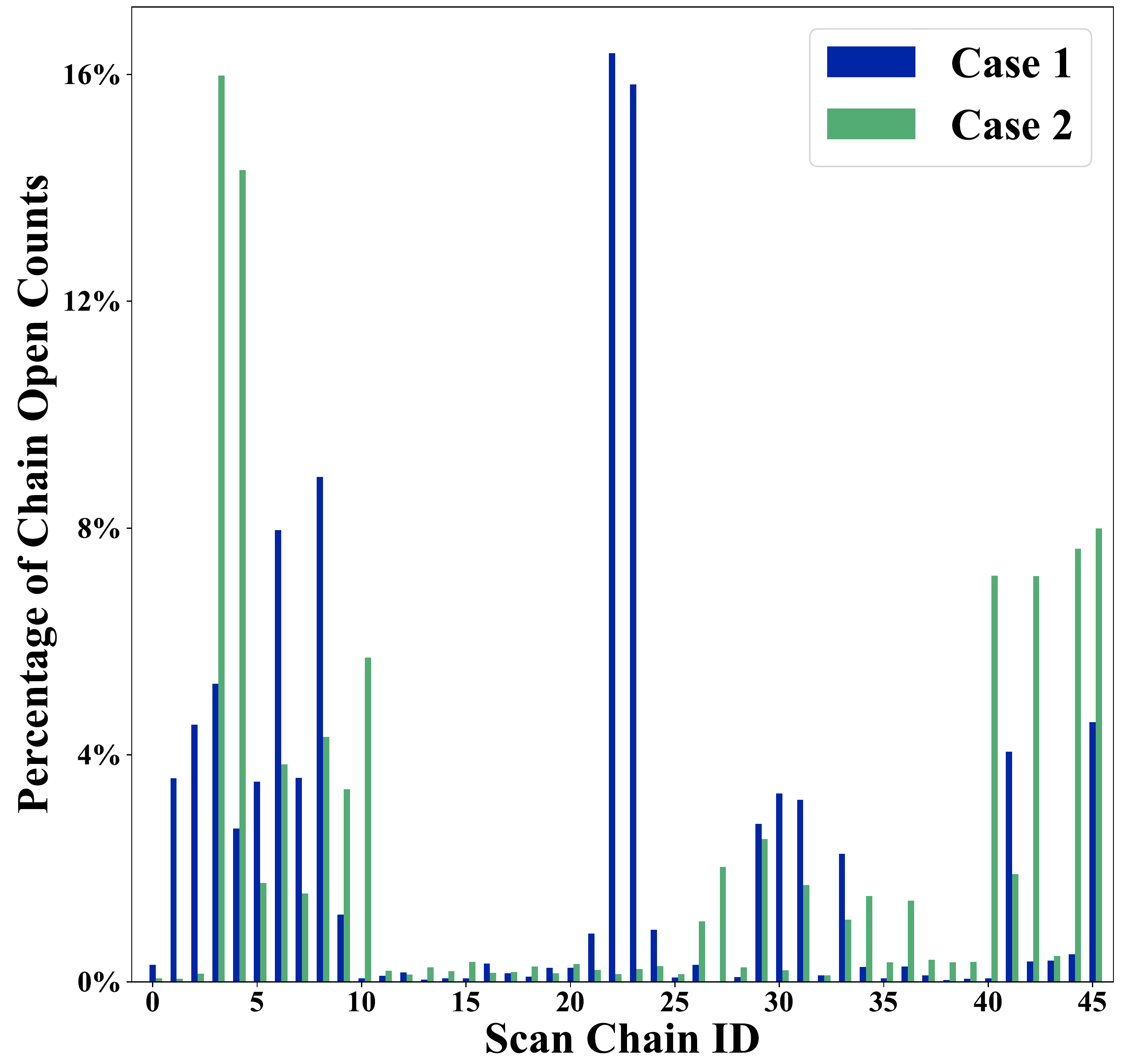}
    \caption{The Distribution of Activated Scan Chains}
    \label{fig:scanchain}
\end{figure}

Although such XORNet design is simple, it ignores circuit-specific  testability characteristics and can be sub-optimal due to this fact. For example, the design of circuit is non-uniform, resulting in uneven distribution of scan chain usage. we analyze two test cases with the same number of scan chains. As shown in the Fig.~\ref{fig:scanchain}, high variance of scan chain usage on a certain case is observed. As for the Case 1, some scan chains can be frequently used, and consume around $16.0\%$ of total specified chain counts, while some other scan chains are activated for less than $1.0\%$ times. What's more, the distribution of specified scan chain counts for two cases varies significantly. To be specific, the usage peak value appears on scan chain \#22 and \#23 for the Case 1, but those scan chains are merely used in Case 2. In the following discussion, the distributions of scan chain usage are regarded as one form of circuit testability properties. 

In general, different circuits maintain different testability properties, which implies a unitary XORNet design is unsatisfactory. In this case, we need different control strengths for different scan chains. If a sufficient number of in-distribution scan chains sampled from a specific circuit are available, then in principle, \textit{Learning-based} approach should be able to extract the underlying patterns and produce a better XORNet structure for this circuit.

Due to the logical properties of XOR operation, XORNet is neither differentiable, nor is its coding ability directly related to the number of control bits. Consequently, conventional optimization methods cannot be directly applied to XORNet design. For example, the gradient descent method cannot deal with the XOR encoding optimization problem. In this paper, we propose an evolutionary-learning-guided approach to search XORNet structure, which is elaborated in Section~\ref{sec:ga}.

\section{Genetic Algorithm for XORNet}
\label{sec:ga}
To realize the idea of customizing XORNets for different circuits in practice, we take the idea of genetic algorithm, one of the evolutionary learning approaches, with the following two reasons:
\begin{itemize}
    \item One row of XORNet matrix $\mathbf{A}$ corresponds to the XOR operand settings for one scan chain. Different rows are independent and do not interfere with each other. Consequently we can modify XORNet at the row level. If two XORNets are given, we can randomly exchange rows of two matrices and obtain a new XORNet. Such procedure can be modeled as the \textit{crossover} operation in GA.
    \item For the $i^{th}$ scan chain, its control bits are determined by entries being $1$ of $\mathbf{a}_i$ ($\mathbf{a}_i$ is the $i^{th}$ row of XORNet matrix $\mathbf{A}$). To explore different setting of control bits for the scan chain, one can flip the entries of $\mathbf{a}_i$. Such entry-wise operation can be regarded as one variation of \textit{mutation} in GA.
\end{itemize}

\begin{figure}[t]
    \centering
    \includegraphics[width=\linewidth]{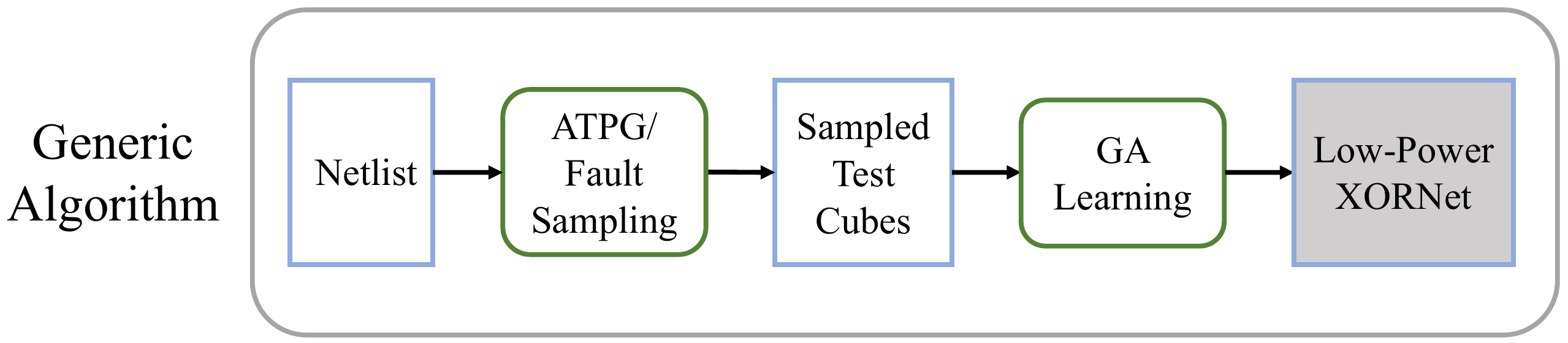}
    \caption{Overview of GA Pipeline.}
    \label{fig:overview}
\end{figure}

To this end, we propose to use GA for searching better XORNet structure in low power controller design. Fig.~\ref{fig:overview} shows the overview of the GA pipeline and the GA learning procedure is illustrated in Algorithm~\ref{algo:ga}. We elaborate the GA for XORNet-based low power controller design in the following sub-sections.

\begin{algorithm}[t]
\caption{GA Learning for Low-Power Test Compression}
\label{algo:ga}
\KwIn{Sampled Test Cubes $\mathbf{X}$, \# of Scan Chain $N$, \# of Control Bits $M$, \# of Individuals $Size\_Pop$, \# of Parents $Size\_Parents$, \# of Parents $Size\_Children$, \# of generations $Size\_Gen$, Mutation ratio $\gamma$, Coefficient $\lambda$.}
\KwOut{XORNet $\mathbf{A}$ searched by GA.}
\tcc{Initialize Population}
$\mathbf{\{\mathbf{A}\}} = \text{InitPop}(N, M, Size\_Pop)$\;
$i = 1$\;
\tcc{Define Fitness Function}
$\ell = \text{UE} + \lambda \times \overline{\text{SCA}}$\;
\While{$\text{Not Convergence or } i \leq Size\_Gen$}{
\tcc{Fitness Calculation}
$\mathbf{fitness} = \ell(\{\mathbf{A}\}, \mathcal{X})$\;
\tcc{Select Parents to Mating Pool}
$\{\textbf{A}_{parent}\} = top(\{\mathbf{A}\}, \mathbf{fitness}, Size\_Parents)$\;
\tcc{CrossOver}
$\{\textbf{A}_{child}\} = CrossOver(\{\textbf{A}_{parent}\}, Size\_Children)$\;
\tcc{Mutation}
$\{\textbf{A}_{child}\} = Mutation(\{\textbf{A}_{child}\}, \gamma)$\;
\tcc{New Population}
$\{\mathbf{A}\} = \{\textbf{A}_{parent}\} + \{\textbf{A}_{child}\}$\;
$i++$\;
}
\tcc{Best XORNet}
$\mathbf{A} = top(\{\mathbf{A}\}, \mathbf{fitness}, 1)$\;
\Return $\mathbf{A}$;
\end{algorithm}

\subsection{Data Preparation}
To enable GA learning, we have to prepare a pool of test cubes $\mathbf{X}$, which approximates the distribution of test cubes in ATPG pipeline. As the XORNet structure is to be determined, the final circuit netlist is not available during the GA training stage, hence we cannot directly run ATPG and sample the exact test tubes. As the low-power controller occupies only a small part of the final circuit, we consider applying \textit{fault sampling} on the original circuit netlist. Specifically, we randomly select a small subset of faults (less than $4.0\%$ of the total faults) and run ATPG, extract their scan chain usage information, and use them as the training dataset for GA learning. The key observation is that a small portion of faults are already sufficient to reflect the statistic of scan chain usage and is able to help GA search a better XORNet structure than conventional XORNet.


\subsection{GA Specifications}

\subsubsection{Population Initialization}
The population consists of multiple XORNet set $\{\mathbf{A}\}$, wherein each XORNet matrix $\mathbf{A}_i \in \{0, 1\}^{N\times M}$ represents an individual solution. 
Because the number of $1$'s directly affects the hardware overhead of XORNet (The more 1's, the larger the area), we restrict the sparsity of $\mathbf{A}$ so that on average there are $3$ control bits being the operands for each scan chain. In other words, $3$ entries of $\mathbf{a}_i$ are $1$'s and the remaining entries are set as $0$. Through this initialization method, we ensure that the area of XORNet searched by the GA is comparable to the one by the previous works.

The most significant factor affecting the search space is the size of population. With more individual solutions in the population, we have more diversity in the searching space. But with too many individuals the convergence speed may be slow. In our experiments, we observe that setting the population size as $40$ is enough to get a desired results across different test circuits.

\subsubsection{Fitness Function for Low-Power Test Design}
The objective of GA learning has two-fold: Firstly, we want to improve the XORNet encoding capacity so that it can merge more test cubes during ATPG. From this aspect, we use $UE$ (defined in Eq.~\ref{eq:UE}) as indicators, which is the number of failure cases of XORNet encoding and validation of the power requirement. Note that in order to efficiently estimate the power consumption, instead of the transition power, we use $SCA$ defined in Eq.~\ref{eq:powe}. Secondly, except for the use-defined power limit, we also want the average power during test to be as low as possible. Thus, $\overline{SCAE}$ (defined in Eq.~\ref{eq:powermean}) should be minimized. Based on the above considerations, we set the fitness function as the sum of $UE$ and $\overline{SCA}$, and use GA learning to minimize such fitness function, as shown in Eq.~\ref{eq:fitness}. In our experiments, we set coefficient $\lambda$ as $100$ to scale up the range of $\overline{SCA}$ from $(0, 1)$ to $(0, 100)$. 
\begin{equation}
    Fitness\ Function = UE + \lambda \times \overline{SCA}
    \label{eq:fitness}
\end{equation}

\subsubsection{Mating}
To select the individuals as parents to generate children (offspring), we evaluate each individual on test cube set $\mathbf{X}$ and put the ones with low fitness into the mating pool. The utility function $top()$ (line 6 in Algorithm~\ref{algo:ga}) ranks the fitness of individuals from the 
lowest to the highest and selects the top XORNets with the size of $ Size\_Parents$ ($5$ in our experiments) accordingly. The set $\{\mathbf{A}_{parent}\}$ being selected is the mating pool for generating children.

\subsubsection{Crossover}
As discussed before, each row of individuals is independent for XORNet decoding. Consequently, To re-combine two XORNets, one feasible way is to do crossover operation at the row level. Specifically, given two $i^{th}$ row vectors $\mathbf{a}_i^{p1}$ and $\mathbf{a}_i^{p2}$ (the superscript indicates the identity of parent individual) from the parent $\mathbf{A}^{p1}$ and the parent $\mathbf{A}^{p2}$, one of then is randomly picked and designated as $\mathbf{a}_i^c$, the $i^{th}$ row of children XORNet $\mathbf{A}^c$:

\begin{equation}
    \text{Crossover: }\mathbf{a}_i^c = RandomPick(\mathbf{a}_i^{p1}, \mathbf{a}_i^{p2})
\end{equation}

\begin{figure}[t]
    \centering
    \includegraphics[width=\linewidth]{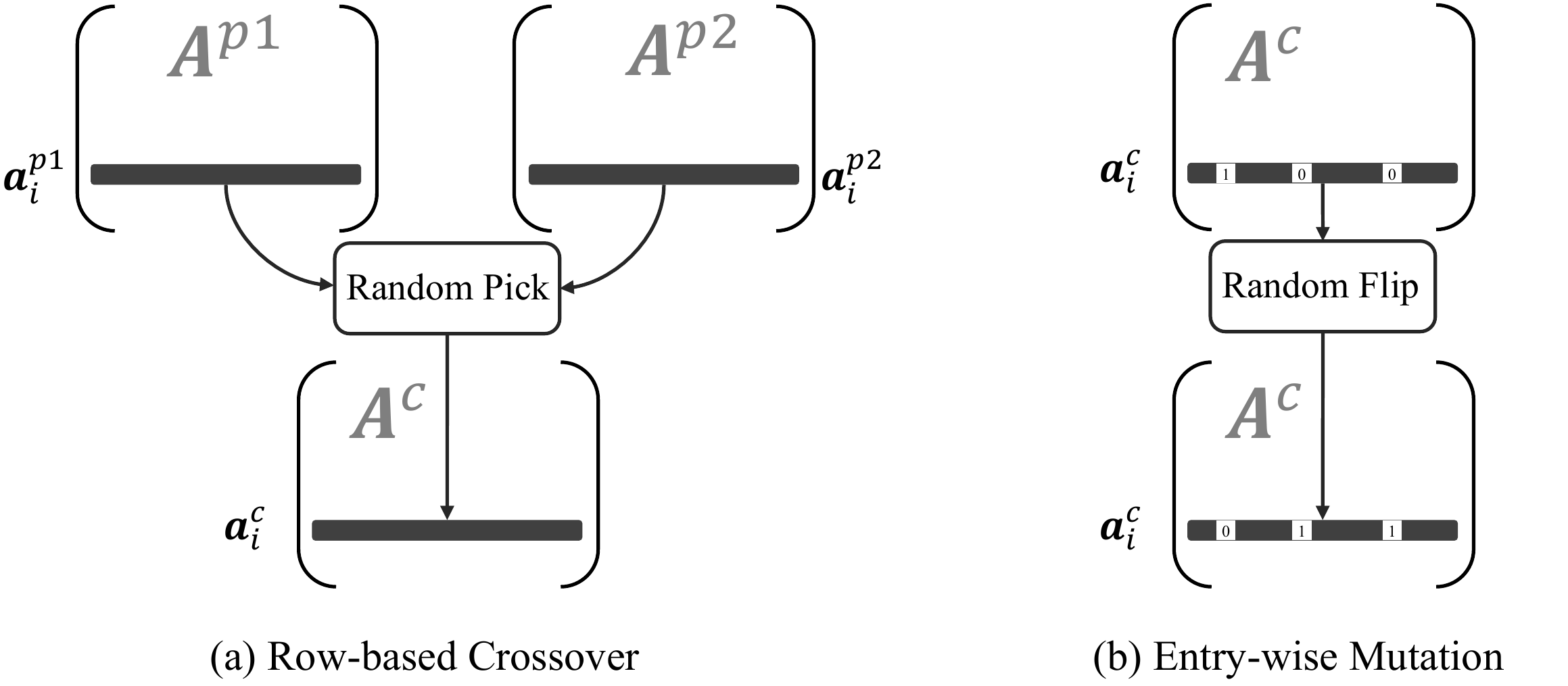}
    \caption{GA Operations}
    \label{fig:gaop}
\end{figure}

The illustration of crossover is shown in Fig.~\ref{fig:gaop}(a). Such crossover results in the rows of $\mathbf{A}^c$ either from $\mathbf{A}_i^{p1}$ or from $\mathbf{A}_i^{p2}$. It gives the chance that the children inherit good XORNet connection settings from both parents and perform better than parents.

\subsubsection{Mutation}
In order to get better solutions, we use mutation to slightly alter the obtained children solutions. As the representation of solution is all binary, bit-flip fits well to XORNet mutation. As shown in Fig.~\ref{fig:gaop}(b), given the $i^{th}$ row of children XORNet, we select the positions of elements to be flipped with the probability of $\gamma$ (set as $0.05$ in our experiments):
\begin{equation}
    \text{Mutation: }\mathbf{a}_i = RandomFlip(\mathbf{a}_i, \gamma)
\end{equation}

After crossover and mutation, we combine the parent set $\{\mathbf{A}_{parent}\}$ and the children set $\{\mathbf{A}_{child}\}$ together as the new generation. Then GA steps into the next evolution iteration. The learning stops when the fitness value of generations is converged or GA runs for predefined $Size\_Gen$ generations. At last, the XORNet with the lowest fitness value is selected as the final output (line 12 in Algorithm~\ref{algo:ga}).

\subsection{Low-Power Requirements}
So far, we discuss a stand-alone XORNet as the decoder. The implicit power limit, in this case, is that the decoded test stimuli toggles 25.0\% test cells randomly on average, as the probability of output of XOR gate being $1$ is $50.0\%$. When users require more stringent power consumption limit, previous works add AND gates to control the probability of scan chain being activated. To be specific, a XORNet $\mathbf{A}$ and a ANDNet $\mathbf{B}$ are combined together so that two outputs from XOR gates are ANDed to get the final output. The proposed GA scheme can be easily generalized to these cases, wherein we fix the ANDNet $\mathbf{B}$ and let GA search for XORNet $\mathbf{A}$. 

\section{Experiments}
\label{sec:exp}

\subsection{Circuit Datasets and Baselines}

\begin{table}[]
\setlength{\tabcolsep}{0.8mm}{
\centering
\caption{Testing circuit characteristics (CBC: Control Bit Counts, PC: Pattern Counts, TC: Test Coverage)}\label{TAB:Circuit}
\begin{tabular}{@{}llll|lll|lll@{}}
\toprule
        &         &         &        & \multicolumn{3}{c}{$R$=50.0\%} & \multicolumn{3}{c}{$R$=30.0\%} \\ 
ID & \# Gate  & \# Chain & \# Cell & CBC   & PC      & TC       & CBC   & PC      & TC       \\ \midrule
C1      & 86,917  & 33      & 328    & 9     & 1,263   & 98.57\%  & 16    & 1,323   & 97.97\%  \\
C2      & 121,259 & 46      & 358    & 11    & 2,260   & 96.01\%  & 23    & 2,539   & 95.82\%  \\
C3      & 782,934 & 81      & 329    & 20    & 10,893  & 97.05\%  & 40    & 13,426  & 96.91\%  \\
C4      & 54,930  & 100     & 360    & 25    & 5,171   & 94.79\%  & 50    & 2,027   & 88.35\%  \\
C5      & 107,438 & 128     & 328    & 25    & 1,393   & 99.47\%  & 64    & 1,154   & 97.80\%  \\
C6      & 136,112 & 380     & 40     & 76    & 3,850   & 93.80\%  & 190   & 3,876   & 94.05\%  \\
C7      & 91,115  & 400     & 31     & 80    & 2,471   & 98.10\%  & 200   & 2,540   & 98.00\%  \\
C8      & 104,155 & 413     & 330    & 82    & 661     & 99.56\%  & 82    & 656     & 99.51\%  \\
C9      & 86,915  & 500     & 25     & 100   & 2,451   & 98.59\%  & 250   & 2,558   & 98.49\%  \\
C10     & 136,872 & 760     & 20     & 114   & 3,952   & 93.66\%  & 152   & 3,952   & 93.66\%  \\ \bottomrule
\end{tabular}
}
\end{table}

We compare the proposed testability-aware XORNet design searched by GA with the standard XORNet design adopted by commercial tools on 10 industrial circuit designs, ranging in size from $54K$ to $782K$ gates. Table~\ref{TAB:Circuit} summarizes the circuit characteristics including the number of gates (\# Gate), the number of scan chains (\# Chain), the maximum number of scan cells for each scan chain (\# Cell). 
Besides, we consider the test compression with two low power requirements: maximum transition rates $R=50\%$ and $R=30\%$, which represents the peak shift cycle power limit. When the shift cycle power limit is $50.0\%$, the one-level of XORNet structure is deployed. When the shift cycle power limit is $30.0\%$, the two-level XORNet structure is deployed. 

With two different transition rates, two types of conventional XORNet are generated as baselines. In our experiments, we use the number of scan chains as the random seed to generate the XORNets. Therefore, two different circuits with the same number of scan chains share the same XORNet-based low power controller design.

To detect multiple faults with a single test pattern, we consider the incremental merging scheme. 
To be specific, the ATPG procedure generates some test cubes, and some eligible test cubes are incrementally merged into a test pattern to cover other faults as much as possible without losing any test coverage. Such procedure is executed only if the merging pattern activates scan chains less than the maximum SCA. 

As shown in the second and third columns in Table~\ref{TAB:Circuit}, we report the control bit counts (CBC) of test compression structures, ATPG pattern counts (PC) and test coverage (TC) when the conventional XORNet is used as the low power controller.

\subsection{GA Hyper-parameters and Fitness Function Convergence}
We try different hyper-parameter settings for GA and observe that as long as the population size is set to be relatively large (\textit{i.e.}, $>20$), the fitness function will easily be converged. Consequently, the GA hyper-parameters for experimental results shown in this section are all set as the same as follows: number of generations $Size\_Gen=20$, population size  $Size\_Pop=40$, number of parents for mating pool $Size\_Parents=5$, number of children $Size\_Children=25$, mutation ratio $\gamma=0.05$, coefficient $\lambda=100$.

\begin{figure}[t]
    \centering
    \includegraphics[width=0.85\linewidth]{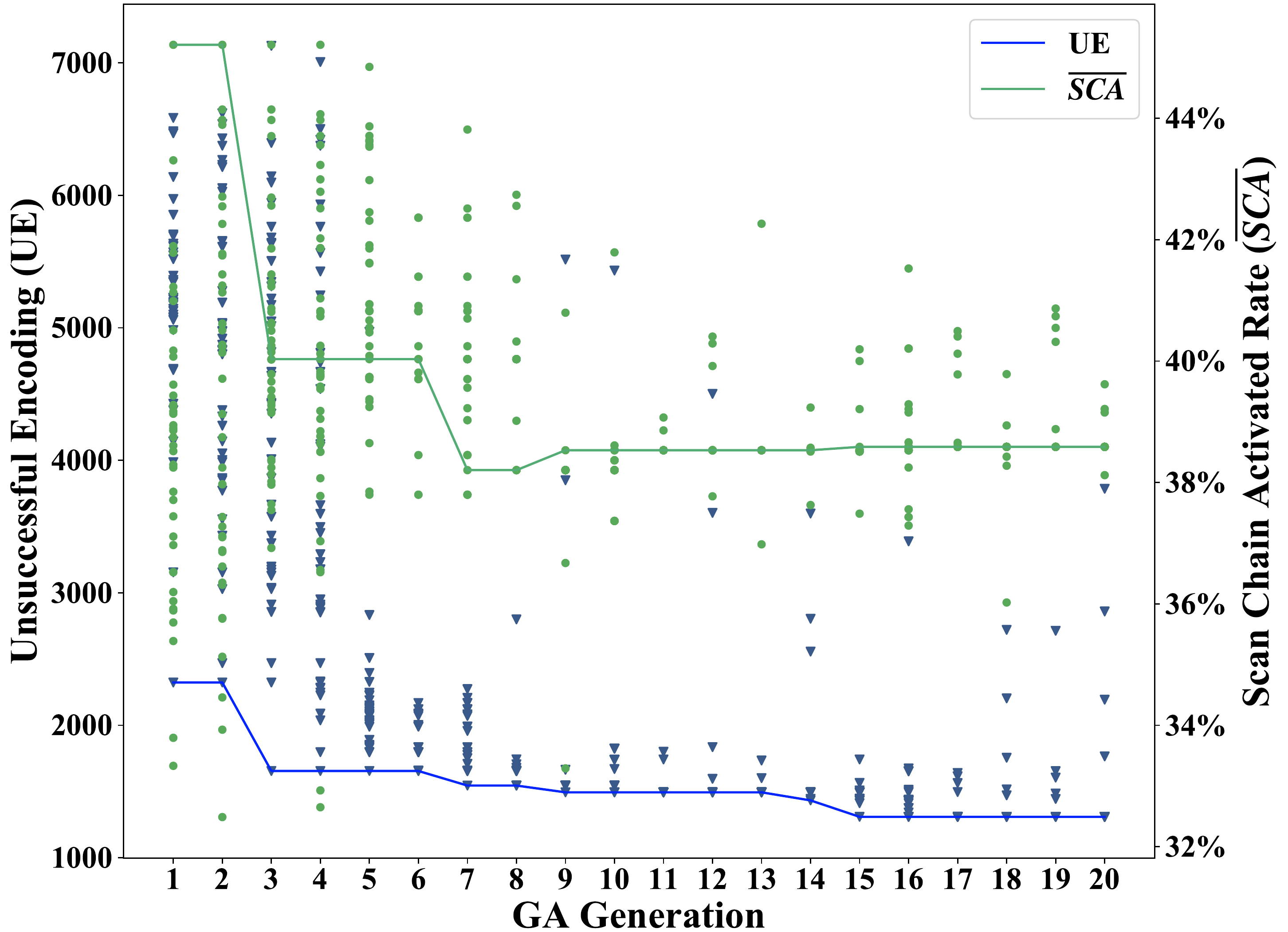}
    \caption{Tracing of the Fitness Function during GA Evolution}
    \label{fig:fitness}
\end{figure}

In Fig.~\ref{fig:fitness}, we sample one circuit (C1) and trace the fitness function values during evolution. Note that a similar trend of fitness value on other circuits is observed. As the fitness function has two components $UE$ (blue dots) and $\overline{SCA}$ (green dots), we show them independently in the figure. In each generation ($x$-axis), the fitness values ($y$-axis) for all $40$ XORNet solutions are demonstrated, while the fitness values of the best individuals for all generations are connected as line charts (See the blue line and green line in the figure). As can be observed in this figure, both $UE$ and $\overline{SCA}$ go down and converge within $20$ generations. Because we prioritize $UE$ optimization in our fitness function, it can be seen in the figure that the final XORNet solution has a minimum $UE$, and may have sub-optimal $\overline{SCA}$.

\subsection{Effectiveness of GA on XORNet Design}
As demonstrated in Fig.~\ref{fig:scanchain}, different circuits exhibit different testability properties. The testability-aware low power XORNet takes these properties into account by applying genetic algorithm on sampled test cubes, thus improve the encoding capability for a specific circuit. In this subsection, we demonstrate the effectiveness of GA in terms of the test coverage, testing time and the number of control bits.

\subsubsection{Improve Test Coverage}
The low power controller encoding process is trying to find a qualified solution of linear equations that all pre-specific gating signals should be assigned the correct value after decoding. At the same time, the number of activated scan chains by gating signals cannot exceed the maximum SCA, otherwise, the test cube will be counted as HP and be dropped during ATPG. 

Table~\ref{TAB:Res:Coverage} compares GA-guided XORNet with baselines w.r.t the test coverage. We list four circuits (C1-C4) with HP after coupling the low power test compression structure, while the other cases (C5-C10) are ignored because they have no HP. When the maximum transition rate $R=50.0\%$, the GA learned encoding XORNet can reduce almost all HP and improve by $2.11\%$ test coverage at most. When $R=30.0\%$, the GA-guided XORNet improves test coverage up to $0.63\%$. Comparing with the conventional XORNet, the average TC on $4$ circuits increase by $1.25\%$ and $0.43\%$ under two different low power limits, which is significant.

Overall, compared with the conventional low power test compression structure, with embedded testability information, the GA-guided low power controller can avoid activating more than necessary scan chains for the test cubes which are regarded as HP on the former structure. Therefore, the GA learned controller can improve the test coverage. 

\begin{table}[t]
\centering
\caption{ATPG Test Coverage}\label{TAB:Res:Coverage}
\begin{tabular}{@{}lclll@{}}
\toprule
ID               & \multicolumn{1}{l}{$R$}         & TC Baseline       & TC        & Improve         \\ \midrule
C1               & \multirow{4}{*}{\rotatebox{90}{50.0\%}}         & 98.57\%   & 98.63\%   & 0.06\%          \\
C2               &                               & 96.01\%   & 97.26\%   & 1.25\%          \\
C3               &                               & 97.05\%   & 99.16\%   & 2.11\%          \\
C4               &                               & 94.79\%   & 96.37\%   & 1.58\%          \\ \midrule
\textbf{Average} & \multicolumn{1}{l}{\textbf{}} & \textbf{} & \textbf{} & \textbf{1.25\%} \\ \midrule
C1               & \multirow{4}{*}{\rotatebox{90}{30.0\%}}         & 97.97\%   & 98.50\%   & 0.53\%          \\
C2               &                               & 95.82\%   & 96.16\%   & 0.34\%          \\
C3               &                               & 96.91\%   & 97.54\%   & 0.63\%          \\
C4               &                               & 88.35\%   & 88.57\%   & 0.22\%          \\ \midrule
\textbf{Average} & \multicolumn{1}{l}{\textbf{}} & \textbf{} & \textbf{} & \textbf{0.43\%} \\ \bottomrule
\end{tabular}
\end{table}

\subsubsection{Reduce Testing Time under The Same Coverage}
The design of test compression structure targets reducing test data volume and testing time. In this testability-aware low power test compression structure, the controller keeps the current control bits during shifting each test pattern. For each pattern, the load values should be shifted in bits by bits. Thus, the \textit{total testing cycle} indicating testing time is noted as:
\begin{equation}
    Total\ Testing\ Cycle = (\frac{\text{CBC}+D}{C_{in}} + \text{\# Cell}) \times \text{PC} \label{equ:cycle}
\end{equation}
The constant $D$ represents the number of configuration bits for test pattern decompression, which is determined by circuit size. $C_{in}$ is the number of input channels of test compression structure, and all circuits are configured with the same $C_{in} = 1$. The CBC, \# Cell and PC are shown in Table~\ref{TAB:Circuit}. The low power controller needs $(CBC+D/C_{in})$ cycles to configure the control register and testing register and $\# Cell$ cycles to shift in the current test pattern. 

To ensure that the same volume of faults are covered when testing the circuits with the conventional and testability-aware low power compression structures, the new test cubes will be generated until the ATPG process reaches the target test coverage. The target test coverages for each circuit with two maximum transition rates are set as the two columns under TC in Table~\ref{TAB:Circuit} respectively. 
Table~\ref{TAB:Res:PC} shows that the testability-aware low power test compression structure can reduce at most $45.23\%$ and $15.16\%$ pattern counts with $R=50.0\%$ and $R=30.0\%$. According to the Eq.~\ref{equ:cycle}, without modifying the low power controller structure, the variation of total testing cycles is only related to the pattern counts, \textit{i.e.} reduce $45.23\%$ and $15.16\%$ testing cycles. On average, the evolutionary testing compression approach save $10.18\%$ and $6.09\%$ testing time when the maximum transition rates are $50.0\%$ and $30.0\%$, respectively. 

From the above, the GA-guided low power controller can reduce the number of activated scan chains for most test cubes. Thus, the incremental merging process is carried out more times until reaching the maximum SCA. More test cubes can be merged into a test pattern and the pattern counts (PC) can be reduced. 

\begin{table}[t]
\centering
\caption{ATPG Pattern Counts}\label{TAB:Res:PC}
\begin{threeparttable}
\begin{tabular}{@{}lclllll@{}}
\toprule
ID               & \multicolumn{1}{l}{$R$}       & PC*       & PC        & Cycles*   & Cycles     & Reduction        \\ \midrule
C1               & \multirow{10}{*}{\rotatebox{90}{50.0\%}}        & 1,263     & 1,192     & 440,787   & 416,008   & 5.62\%           \\
C2               &                               & 2,260     & 1,856     & 861,060   & 707,136   & 17.88\%          \\
C3               &                               & 10,893    & 8,320     & 3,954,159 & 3,020,160 & 23.62\%          \\
C4               &                               & 5,171     & 5,094     & 2,073,571 & 2,042,694 & 1.49\%           \\
C5               &                               & 1,393     & 1,387     & 526,554   & 524,286   & 0.43\%           \\
C6               &                               & 3,850     & 3,646     & 739,200   & 700,032   & 5.30\%           \\
C7               &                               & 2,471     & 2,444     & 471,961   & 466,804   & 1.09\%           \\
C8               &                               & 661       & 362       & 327,195   & 179,190   & 45.23\%          \\
C9               &                               & 2,451     & 2,427     & 551,475   & 546,075   & 0.98\%           \\
C10              &                               & 3,952     & 3,944     & 980,096   & 978,112   & 0.20\%           \\ \midrule
\textbf{Average} & \multicolumn{1}{l}{\textbf{}} & \textbf{} & \textbf{} & \textbf{} & \textbf{} & \textbf{10.18\%} \\ \midrule
C1               & \multirow{10}{*}{\rotatebox{90}{30.0\%}}        & 1,323     & 1,145     & 470,988   & 407,620   & 13.45\%          \\
C2               &                               & 2,539     & 2,516     & 997,827   & 988,788   & 0.91\%           \\
C3               &                               & 13,426    & 11,391    & 5,142,158 & 4,362,753 & 15.16\%          \\
C4               &                               & 2,027     & 1,994     & 863,502   & 849,444   & 1.63\%           \\
C5               &                               & 1,154     & 1,018     & 481,218   & 424,506   & 11.79\%          \\
C6               &                               & 3,876     & 3,772     & 1,186,056 & 1,154,232 & 2.68\%           \\
C7               &                               & 2,540     & 2,423     & 789,940   & 753,553   & 4.61\%           \\
C8               &                               & 656       & 654       & 324,720   & 323,730   & 0.30\%           \\
C9               &                               & 2,558     & 2,367     & 959,250   & 887,625   & 7.47\%           \\
C10              &                               & 3,952     & 3,838     & 1,130,272 & 1,097,668 & 2.88\%           \\ \midrule
\textbf{Average} & \multicolumn{1}{l}{\textbf{}} & \textbf{} & \textbf{} & \textbf{} & \textbf{} & \textbf{6.09\%}  \\ \bottomrule
\end{tabular}
\begin{tablenotes}
  \footnotesize
  \item[*] Baseline: Test with the conventional XORNet
\end{tablenotes}
\end{threeparttable}
\end{table}

\subsubsection{Reduce Control Bit Counts to Reduce Testing Time}
Different from the conventional XORNets, as XORNets generated by GA utilizes testability properties, the latter can use fewer control bits to reach the same test coverage. We try to reduce the control bit counts to save testing cycles further. To explore the influence of  CBC reduction on test coverage and XORNet encoding capability, we take the cases with relatively high TC and more \# Chain as examples (\textit{i.e.} C8 and C9) in Table~\ref{TAB:Res:EBCReduce}. The target test coverage for C8 and C9 are set to $99.56\%$ and $98.57\%$ respectively, and the minimum control bit counts have been reduced by $75.61\%$ and $70\%$, which means the testability-aware network only takes about $30\%$ of original control bit counts to achieve the same coding capability with the randomly connected XOR encoding network. It should be noted that when CBC is excessively reduced, the test coverage cannot meet the target test coverage. Therefore, the minimum CBCs for C8 and C9 are $20$ and $30$ respectively. 

\begin{figure}
    \centering
    \includegraphics[width=0.7\linewidth]{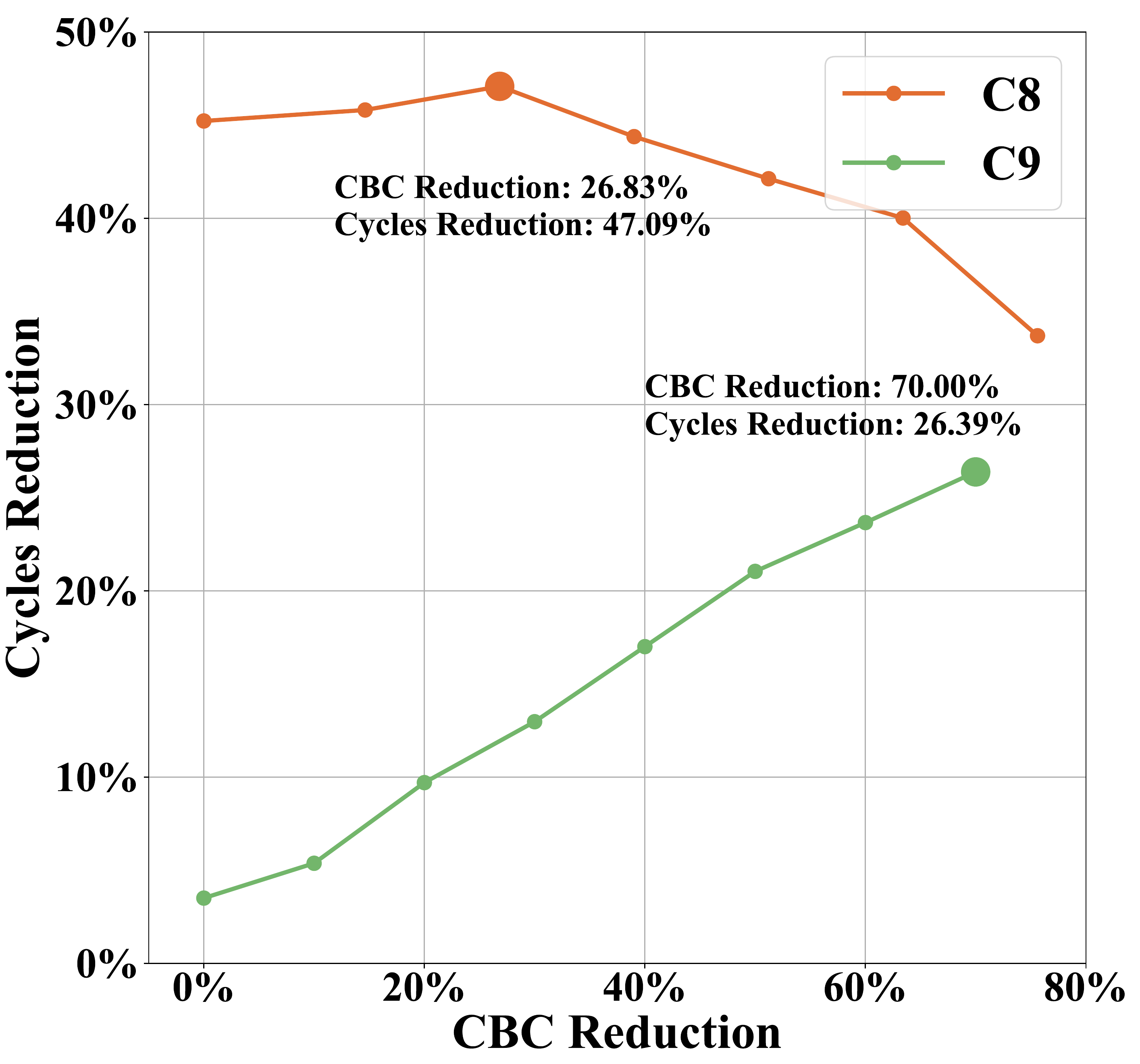}
    \caption{Testing Cycles Reduction with CBC}
    \label{fig:EBCReduction}
\end{figure}

However, the total testing cycle is determined by pattern counts, the longest scan chain and the circuit structure, as the cycles metrics Eq.~\ref{equ:cycle}. By restricting the number of control bits, the XORNet encoding capability decreases and fewer test cubes are merged into a pattern, so that it will lead to the increase of pattern counts to consume testing time. For example, if the CBC of C8 is reduced by $75.61\%$ (from $82$ to $20$), the test cycle reduction does not reach the optimal value. We observe that the XORNet with $26.83\%$ CBC reduction has minimal testing cycles, as shown in Figure~\ref{fig:EBCReduction}.

It is not recommended that reducing control bits for circuits that have few scan chains and CBC, because the reduction of CBC compression structures may activate more percentage of scan chains so that exceed the maximum SCA. Restricting control bit strategy is more suitable for circuits with more scan chains. Therefore, the circuits (C1-C4) whose CBCs are less than $50$ are not considered. According to the above analysis of C8 and C9, the low power controller with the least CBC may not reach the minimal testing cycles. In the Table~\ref{TAB:Res:EBC}, we generate the testability-aware low power test compression structures for the circuits and configure fewer control bits to further reduce the testing time at most. 
The best results are saving $47.09\%$ and $36.11\%$ testing time by reducing control bits when the maximum transition rate is $50.0\%$ and $30.0\%$,  respectively. 
When $R=50.0\%$, the testability-aware approach can reduce $46.99\%$ CBC to save about $20.78\%$ testing cycles on average. Besides, for more power consumption limited situations $R=30.0\%$, GA can reduce $19.40\%$ testing cycles by decreasing about $35.84\%$ CBC.

\begin{table}[t]
\centering
\caption{Reduce Control Bit Counts}\label{TAB:Res:EBCReduce}
\begin{threeparttable}
\begin{tabular}{@{}cllllll@{}}
\toprule
\multicolumn{1}{l}{ID} & CBC & Reduction & PC   & TC      & Cycles   & Reduction \\ \midrule
\multirow{9}{*}{C8}    & 82\tnote{*}  &  --     & 661  & 99.56\% & 327,195 & --  \\
                       & 82  & 0.00\%    & 362  & 99.56\% & 179,190 & 45.23\%  \\
                       & 70  & 14.63\%   & 367  & 99.56\% & 177,261 & 45.82\%  \\
                       & 60  & 26.83\%   & 366  & 99.56\% & 173,118 & 47.09\%  \\
                       & 50  & 39.02\%   & 393  & 99.56\% & 181,959 & 44.39\%  \\
                       & 40  & 51.22\%   & 418  & 99.56\% & 189,354 & 42.13\%  \\
                       & 30  & 63.41\%   & 443  & 99.56\% & 196,249 & 40.02\%  \\
                       & \textbf{20}  & \textbf{75.61}\%   & \textbf{501}  & \textbf{99.56}\% & \textbf{216,933} & \textbf{33.70}\%  \\
                       & 10  & 87.80\%   & 649  & 99.48\% & 274,527 & 16.10\%  \\ \midrule
\multirow{10}{*}{C9}   & 100\tnote{*} & --      & 2,451 & 98.57\% & 551,475 &  --        \\
                       & 100 & 0.00\%    & 2,365 & 98.57\% & 532,125 & 3.51\%   \\
                       & 90  & 10.00\%   & 2,427 & 98.57\% & 521,805 & 5.38\%   \\
                       & 80  & 20.00\%   & 2,429 & 98.57\% & 497,945 & 9.71\%   \\
                       & 70  & 30.00\%   & 2,461 & 98.57\% & 479,895 & 12.98\%  \\
                       & 60  & 40.00\%   & 2,474 & 98.57\% & 457,690 & 17.01\%  \\
                       & 50  & 50.00\%   & 2,488 & 98.57\% & 435,400 & 21.05\%  \\
                       & 40  & 60.00\%   & 2,551 & 98.57\% & 420,915 & 23.67\%  \\
                       & \textbf{30}  & \textbf{70.00}\%   & \textbf{2,619} & \textbf{98.57}\% & \textbf{405,945} & \textbf{26.39}\%  \\
                       & 20  & 80.00\%   & 2,691 & 98.44\% & 390,195 & 29.25\%  \\ \bottomrule
\end{tabular}
\begin{tablenotes}
  \footnotesize
  \item[*] Baseline: Test with the conventional XORNet
\end{tablenotes}
\end{threeparttable}
\end{table}

\begin{table*}[th!]
\centering
\caption{Reduce Control Bit Counts}\label{TAB:Res:EBC}
\begin{tabular}{@{}lclll|lll|lll@{}}
\toprule
ID               & \multicolumn{1}{l}{$R$}       & CBC Baseline      & CBC       & Reduction        & PC Baseline       & PC        & Reduction       & Cycles Baseline   & Cycles     & Reduction        \\ \midrule
C5               & \multirow{6}{*}{\rotatebox{90}{50.0\%}}         & 25        & 20        & 20.00\%          & 1,393     & 1,402     & -0.65\%         & 526,554   & 522,946   & 0.69\%           \\
C6               &                               & 76        & 40        & 47.37\%          & 3,850     & 3,648     & 5.25\%          & 739,200   & 569,088   & 23.01\%          \\
C7               &                               & 80        & 60        & 25.00\%          & 2,471     & 2,493     & -0.89\%         & 471,961   & 426,303   & 9.67\%           \\
C8               &                               & 82        & 60        & 26.83\%          & 661       & 366       & 44.63\%         & 327,195   & 173,118    & 47.09\%          \\
C9               &                               & 100       & 30        & 70.00\%          & 2,451     & 2,619     & -6.85\%         & 551,475   & 405,945   & 26.39\%          \\
C10              &                               & 114       & 50        & 56.14\%          & 3,952     & 4,001     & -1.24\%         & 980,096   & 736,184   & 24.89\%          \\ \midrule
\textbf{Average} & \multicolumn{1}{l}{\textbf{}} & \textbf{} & \textbf{} & \textbf{46.99\%} & \textbf{} & \textbf{} & \textbf{4.77\%} & \textbf{} & \textbf{} & \textbf{20.78\%} \\ \midrule
C5               & \multirow{6}{*}{\rotatebox{90}{30.0\%}}         & 64        & 55        & 14.06\%          & 1,154     & 946       & 18.02\%         & 481,218   & 385,968   & 19.79\%          \\
C6               &                               & 190       & 110       & 42.11\%          & 3,876     & 3,793     & 2.14\%          & 1,186,056 & 857,218   & 27.73\%          \\
C7               &                               & 200       & 160       & 20.00\%          & 2,540     & 2,556     & -0.63\%         & 789,940   & 692,676   & 12.31\%          \\
C8               &                               & 82        & 40        & 51.22\%          & 656       & 683       & -4.12\%         & 324,720   & 309,399   & 4.72\%           \\
C9               &                               & 250       & 100       & 60.00\%          & 2,558     & 2,724     & -6.49\%         & 959,250   & 612,900   & 36.11\%          \\
C10              &                               & 152       & 110       & 27.63\%          & 3,952     & 3,902     & 1.27\%          & 1,130,272 & 952,088   & 15.76\%          \\ \midrule
\textbf{Average} & \multicolumn{1}{l}{\textbf{}} & \textbf{} & \textbf{} & \textbf{35.84\%} & \textbf{} & \textbf{} & \textbf{1.70\%} & \textbf{} & \textbf{} & \textbf{19.40\%} \\ \bottomrule
\end{tabular}
\end{table*}

\subsection{GA-guided XORNet Analysis}

In this section, we explore and analyze why GA-guided XORNet outperforms conventional XORNet from the perspective of control bit connection and scan chain activated frequency during ATPG. 

\begin{figure}[t]
    \centering
    \includegraphics[width=0.75\linewidth]{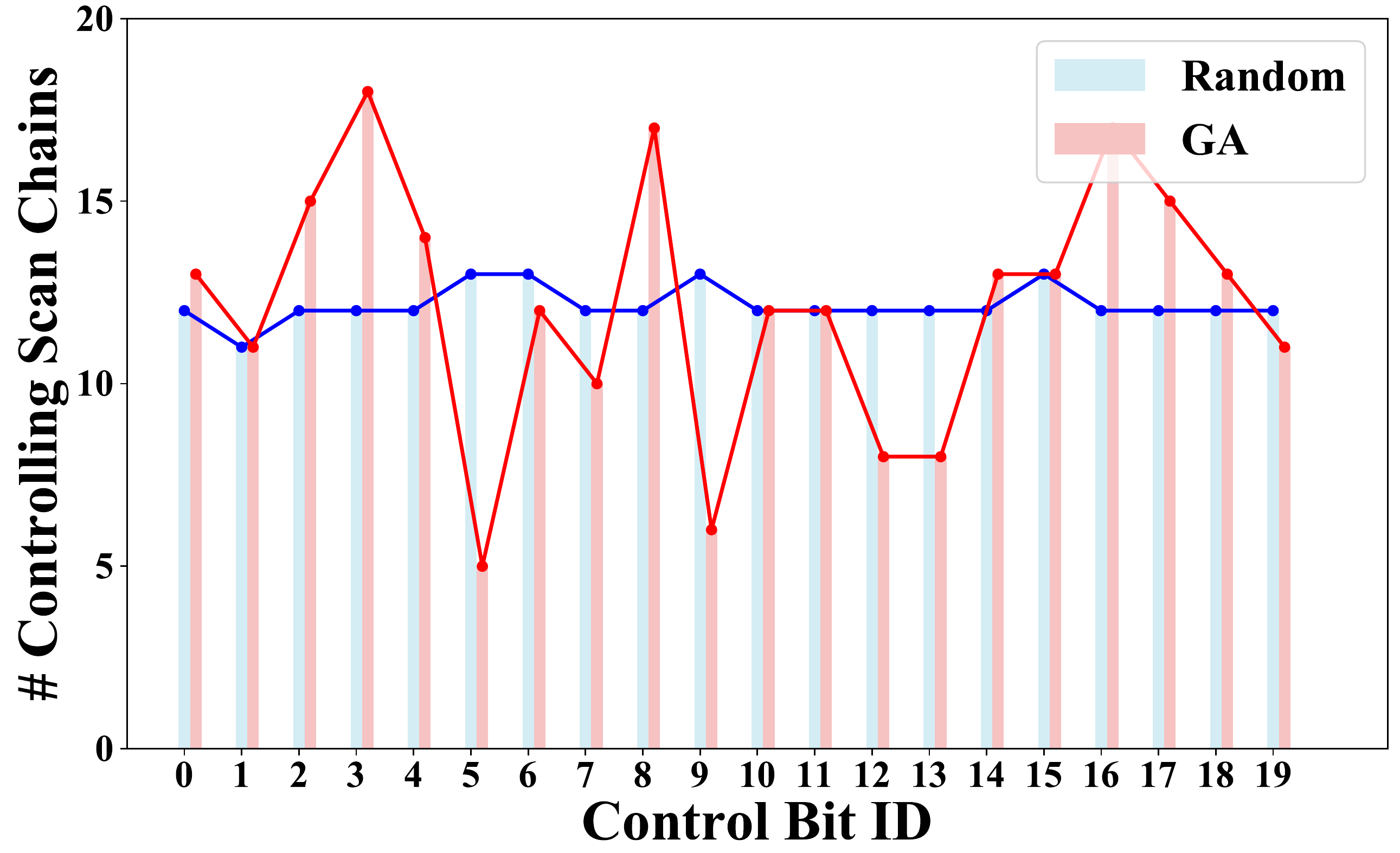}
    \caption{Number of Scan Chains Controller by Each Control Bit}
    \label{fig:CBCControl}
\end{figure}

Firstly, we show the number of scan chains controlled by each control bits for case C3 in Fig.~\ref{fig:CBCControl}. As in the XORNet-based low power controller, the control bits are expanded into final gating signals for scan chains. Usually, the number of control bits is much less than the number of scan chains. In other words, one control bit will participate in the control of multiple scan chains. As can be observed from this figure, the conventional XORNet (the blue line) is configured randomly and each encoding bit controls a similar number of scan chains (\textit{i.e.}, around 12). On the contrary, the GA-guided XORNet (the red line) can differentiate different scan chain (\textit{i.e.} a control bit can control 18 at most and 5 at least) and assign different numbers of control bits for each scan chain. Such results are in line with our expectation as GA-guided XORNet can exert different control strengths on different scan chains.


\begin{figure}[t]
    \centering
    \includegraphics[width=0.9\linewidth]{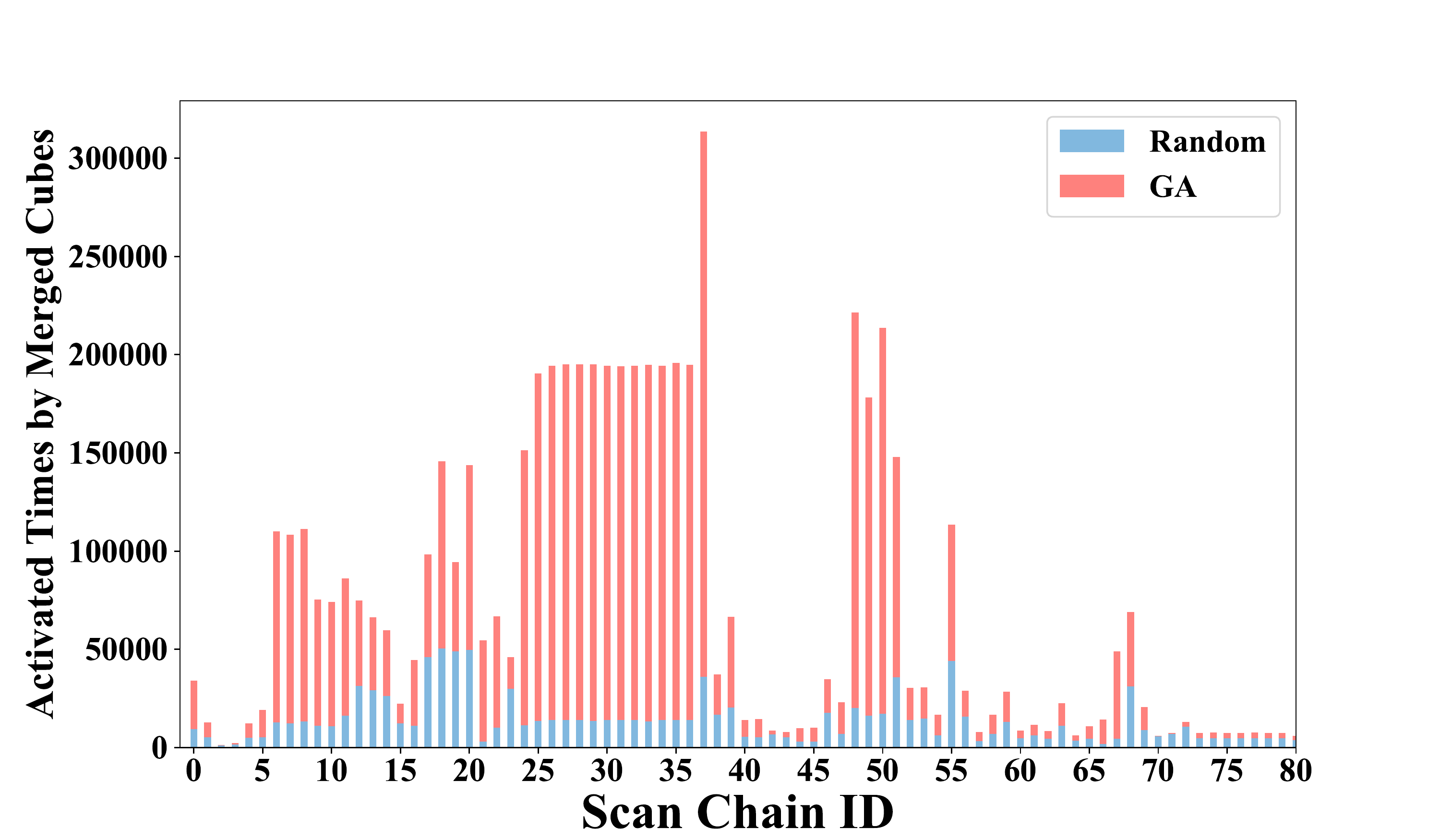}
    \caption{The Activated Times of Each Scan Chains}
    \label{fig:GAImprove}
\end{figure}

Secondly, to show more evidence of better encoding capacity obtained from GA-guided XORNet, we count which scan chain needs to be activated by successfully merged cubes during incremental merging. This indicator directly corresponds to the test coverage and reflects the coding ability of XORNet. The more activated times of scan chains, the more test cubes are generated and merged successfully which are able to detect more faults and improve the test coverage. In order to show the GA-guided XORNet advantage clearly, we take the case (C3) with the largest test coverage improvement as an example. As shown in Figure~\ref{fig:GAImprove}, GA-guided XORNet (the red bar) constantly outperforms the conventional XORNet (the blue bar) regarding the activated times of scan chain during ATPG. The relative improvement is essentially large on some scan chains (\textit{e.g.}, scan chain \#37), which shows the benefit of GA-guided XORNet. Similar improvements are emerging in other cases.




To sum up, GA-guided XORNet is able to extract the testability properties from sampled test cubes, and rearrange the control bit connections so that the encoding capacity outperform conventional XORNet by a significant margin.


\section{Conclusions}
\label{sec:conclusion}
In this paper, we propose to use an evolutionary learning algorithm to improve the performance of low power test compression. By applying the genetic algorithm on sampled test cubes to extract testability information, and further employing XORNet searched by genetic algorithm as the low power controller, the proposed design can achieve better encoding capacity of XORNet.
The empirical results show that the GA-guided XORNet can improve test coverage by up to $2.11\%$ and $0.63\%$ when the maximum transition rate $R$ is $50.0\%$ and $30.0\%$, respectively. 
Meanwhile, the testability-aware low power controller allows merging more test cubes, which effectively decreases the testing time by $10.18\%$ with $R=50.0\%$ and $6.09\%$ with $R=30.0\%$ on average. 
Restricting the number of control bits can further reduce the testing cycles by $20.78\%$ and $19.40\%$ on average when the maximum transition rate is $50.0\%$ and $30.0\%$, respectively. 
In conclusion, compared with the conventional low power controller, the testability-aware controller consisting of GA-guided XORNet can improve the test coverage and reduce testing cycles without loss of fault coverage.


\end{document}